\documentclass{article}
\usepackage{arxiv}
\usepackage[utf8]{inputenc}
\usepackage{graphicx}
\usepackage{amsmath}
\usepackage{amssymb}
\usepackage[table,xcdraw]{xcolor}
\usepackage{rotating}
\usepackage{multirow}
\usepackage{makecell}
\DeclareMathOperator*{\argmax}{arg\,max}
\DeclareMathOperator*{\argmin}{arg\,min}
\usepackage{flushend}

\title{GuideBP: Guiding Backpropagation Through Weaker Pathways of Parallel Logits}

\author {
Bodhisatwa Mandal\\
Dept. of CSE\\
Jadavpur University\\
\texttt{bodhisatwam@gmail.com}
\and
Swarnendu Ghosh \thanks{Swarnendu Ghosh and Nibaran Das are the corresponding authors}\\
Dept. of CSE\\
Jadavpur University\\
\texttt{swarbir@gmail.com}
\and
Teresa Gon\c{c}alves\\
Dept. of Informatics\\
University of Evora\\
\texttt{tcg@uevora.pt}
\and
Paulo Quaresma\\
Dept. of Informatics\\
University of Evora\\
\texttt{pq@uevora.pt}
\and
Mita~Nasipuri\\
Dept. of CSE\\
Jadavpur University\\
\texttt{mita.nasipuri@jadavpuruniversity.in}
\and
Nibaran Das $^*$\\
Dept. of CSE\\
Jadavpur University\\
\texttt{nibaran.das@jadavpuruniversity.in}
}
\begin{document}
\maketitle
\begin{abstract}
Convolutional neural networks often generate multiple logits and use simple techniques like addition or averaging for loss computation. But this allows gradients to be distributed equally among all paths. The proposed approach guides the gradients of backpropagation along weakest concept representations. A weakness scores defines the class specific performance of individual pathways which is then used to create a logit that would guide gradients along the weakest pathways. The proposed approach has been shown to perform better than traditional column merging techniques and can be used in several application scenarios. Not only can the proposed model be used as an efficient technique for training multiple instances of a model parallely, but also CNNs with multiple output branches have been shown to perform better with the proposed upgrade. Various experiments establish the flexibility of the learning technique which is simple yet effective in various multi-objective scenarios both empirically and statistically. 
\end{abstract}



\section{Introduction}
Convolutional Neural Networks(CNNs) have come a long way since their introduction of the LeNet5 in 1998\cite{lenet}. Over the last two decades, several architectures have been proposed to solve different types of computer vision tasks\cite{survey1,survey2}. Throughout the evolution of CNNs, it has been observed that the topology in which the layers of CNNs are arranged has a significant impact on performance\cite{survey3}. In the earlier years alternating convolution and pooling layers were used for feature extraction before passing them to a fully connected layers for mapping the features to the output \cite{lenet,alexnet,caps}. Deeper networks were proposed in subsequent years to improve upon previous models. Convolutional blocks with multiple convolution and batch normalization layers were used for extracting rich features at different scales \cite{vggnet}. With deeper networks, the flow of gradients through long chains of computation became a challenge due to issues like vanishing gradients. In this context, major strides have been made by manipulating the CNN architecture. GoogLeNet\cite{googlenet} used auxiliary classifiers to improve gradients of intermediate layers while ResNet and DenseNet implemented skip connections for better gradient flow through short-cut channels. While standard networks like these are good at learning a set of generalized features, multi-column architectures aim for a much more distributed learning protocol\cite{ciresan,ciresan2} where each column is designed to attend a specific purpose. Multiple columns drawn from different layers of a CNN can be used for mapping features of different scales to the output space. This is commonly used of object detection \cite{yolo,yolov3} where objects of interest can vary in terms of scale in the image. For datasets like handwritten texts where samples adhere to a geometric template, region specific columns have been quite successful\cite{sarkhel2017multi, ukil1, ukil2}. However, in most multi column architectures, the columns are either trained independently or parallelly. While each column specializes in learning specific features, their contribution to the output distribution has no effect on the training process. The most common approach to deal with multi path network involves individual training of each path using separate loss functions or adding or averaging the logits.
\par In the proposed methodology a novel learning technique has been developed which emphasizes on exclusively improving the pathways of multi column networks that contribute adversely to output layer. Unlike other aggregation methods like adding or averaging logits, in this method the gradient isn't equally split along all pathways. The learning model considers multiple output distributions from individual columns and forwards the most uncertain pathways to the prediction layer for guiding the backpropagation algorithm along the weakly learned features. This allows all columns of the networks to perform at par with each other. The proposed learning technique can be used wherever a CNN has multiple branches connecting to the output layer. Various application scenarios have been demonstrated that employs the proposed technique to avoid a greedy approach for column optimizations.
\par The objective of the proposed work is to provide a more efficient way to combine multiple sets of logits generated by multi path convolutional neural networks. More specifically, the major contribution lies in the method of combining the logits and establishing as a superior method over standard techniques like adding or averaging. Since we are focusing on a generic learning mechanism for multi-path convolutional neural networks, the proposed method is not being established as a bench-marking model. As most of the applications demonstrated in the work are built as an upgrade over existing models\cite{resnet, sarkhel2017multi, googlenet, yolov3}, it comes with some of the drawbacks of the back-end models themselves.


\begin{figure*}[]
	\centering
	\includegraphics[width=\textwidth]{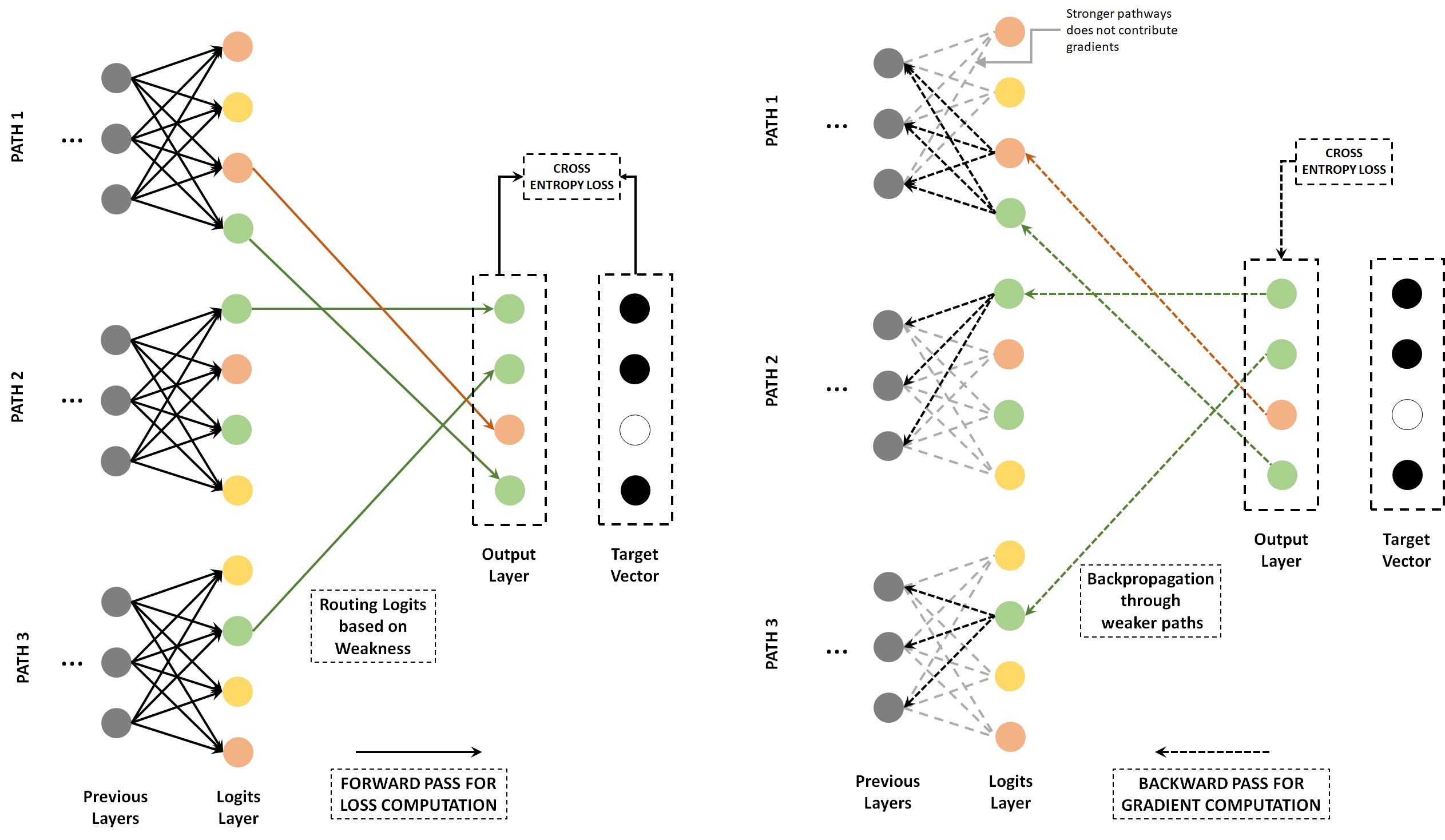}
	\caption{Guiding backpropagation through weaker pathways}
	\label{f:bp}
\end{figure*}

\section{Multi-objective optimizations in CNNs}
Convolutional neural networks are traditionally trained by computing gradients corresponding to trainable parameters that contribute towards a loss function. A stochastic gradient descent based loss minimization technique gradually pushes the weights towards  minima in the loss versus weight space. The loss function has significant impact on several factors that control the performance of the network. While stochastic approaches like gradient descent makes the network prone to issues like local optima and over-fitting, loss functions involving multiple output vectors can negate these issues to a considerable extent as compared to loss functions corresponding to a single output vector. While coping up with multiple objectives a generalized learning paradigm is enforced. In practice, multiple objectives can be enforced at various levels. While some approaches tune differently initialized instances of the same model, other approaches consider things features of multiple scales from different layers, different regions of the feature space and even different regions of the input space.

\subsection{Ensemble of multiple instances of a CNN}
One of the most coarse approach for multi-objective is to take an ensemble of fully functional CNNs. CNNs, by their design, depends on the initialization of the parameters. By taking multiple instances of the same model, we can explore weight space starting from different initial positions. Such models can be either trained individually and combined by ensemble techniques such as max voting, or have their logits summed or averaged before computation of the loss function. Either way, the individual pathways are trained greedily and do not approach to attain the objective jointly.

\subsection{Using auxiliary classifiers to facilitate gradient propagation}
With the expansion of network depth, the chance of gradients vanishing in earlier layers is increasing due to the use of successive products partial derivatives. In networks like GoogLeNet, and Inception Net, the gradients in the earlier layers are boosted using a set of auxiliary classifiers drawn from features of intermediate layers. The purpose of these layers is to introduce an additive component to the gradients of the earlier layers to make up for the drop in the value of the gradients due the depth of the network. Though auxiliary classifiers were optimized according to the same ground-truth as the output layer, however, their learning strategy is greedy. While the gradients in the common pathways were summed, there is no control of the pathways with regards to the strength of the performance. For example, a component of the logits from the auxiliary classifier corresponding to a specific class may be better at contributing to the performance as compared to their equivalent counterpart in the output layer. This can happen due to the fact that all classes do not require the same amount of depth to model the concepts. Excess amount of depth in the network can overfit the training data corresponding to samples of some class.

\subsection{Multi scale region extraction from feature space}
Object detection in the real world is a particularly tricky task because of the variable shape and size of objects in the scene. Popular object detection networks like YOLO thus highly depends on extracting features from multiple scales, to create a scale space pyramid of anchor boxes that can serve as candidates for bounding boxes. The YOLO model divides the feature space of the intermediate layers into grids of regions which are then used for detecting the presence of a bounding box centre. However, these multiple regions only contribute to recommendation of separate candidate boxes. In the proposed work we explore a variation different regions of the feature space has been shown to contribute to a unified learning process while solving a common objective.

\subsection{Multi-column networks operating on different regions of the input}
For datasets of small size and low intra-class variance and inter class variance, it is often seen that operating with specific regions of the input yields better results because different regions of the the sample contributes differently towards the modeling of the output space. The input itself can be divided into different sizes of grids where multi-column convolutional neural networks can work on each of these grid individually. Such multi-column networks are  either trained individually or their logits are summed or averaged to compute loss against the ground truth. Either way, the weaknesses of each branch are untracked and thus leaves room for improvement.


\section{Guiding backpropagation through weaker pathways}

The purpose of the current work is to increase the effect of backpropagation on the more uncertain pathways in a multi-path network. The goal is to stop the pathways that have learnt strong class specific concepts from contributing to the gradient. The weights of the network should give more preference on repairing the weaker concepts modeled in the network. This desired effect of the proposed work is visualized in fig. \ref{f:bp}.
\par The method requires a set of $N$ number of class specific logits, which corresponds to $N$ different pathways. A final set of logits is computed by choosing the most uncertain class specific components across each pathway and combining them. The computation involves calculating the weakness of each component of the predictions given by each path. Finally, a new logit is formed by routing the weakest components among each path. The method is discussed in details in the subsections below.
\subsection{Computing weakness of predictions}
The process requires a $N$ number of $C$ dimensional logits, $N$ is the number of \textbf{outputs} and $C$ is the number of \textbf{classes}. Ideally, the logits(denoted as {$\mathbf{u}$}) of the last layer of a path are supposed to reflect the expectation of the image to belong to a specific class. Since we are dealing with multiple pathways, the logits are passed through a log-softmax function before proceeding to obtain a distribution $\hat{\mathbf{u}}$. 
\begin{equation}
\hat{\mathbf{u}}_{ij} = \log{\frac{\exp({\mathbf{u}_{ij}})}{\sum_{k=1}^C{\exp({\mathbf{u}_{kj}})}}},
\end{equation}
where, $i \in \{1,C\} \text{ and } j\in \{1,N\}$. The log-softmax is calculated to obtain log probabilites of the logits. It is preferred over standard normalization or whitening transforms to enhance maximum expectations and penalize misclassifications severely. 
\par
The weakness of a prediction is represented by the relative value of $\hat{\mathbf{u}}$ among all the pathways. For a target one-hot vector $\mathbf{t}$, the weakness of the log-softmaxed logits can be defined by a function $(W)$ such that
\begin{equation}
W(\hat{\mathbf{u}}_{ij}) = \mathbf{t_i}+(-1)^{(1-\mathbf{t_i})}\times \frac{\hat{\mathbf{u}}_{ij}}{\sum_{m=1}^N{\hat{\mathbf{u}}_{im}}},
\end{equation}
where, $i \in \{1,C\} \text{ and } j\in \{1,N\}$. The weakness value maximizes the pathway with the highest logits corresponding to the negative classes and lowest logits corresponding to the positive class. The intuition behind the idea is that the strength of a classifier  depends on how high valued logits it produces for the positive class and vice versa. Hence, the weakness is defined as the opposite of that. These weakness values corresponding to each of the log-softmaxed logits are used to build the combined logits for channeling the backpropagation algorithm.
\subsection{Building logits with most uncertain pathways}
While building the combined logits the focus should be on choosing the pathway with the highest weakness associated with it for the backpropagation algorithm to tune. The combined logits $\mathbf{v}$ is defined as, 
\begin{equation}
\mathbf{v}_{i} = \argmax_{j}\  W({\hat{\mathbf{u}}_{ij}})
\end{equation}
The utility of composing logits with the weakest components of each pathway, is guiding the backpropagation through the weakest branches first. Throughout the learning phase, the backpropagation will always channelized along the pathway that is producing the most uncertain response among all the different branches, thus improving the global certainty of all the branches.
\begin{figure}[h]
	\centering
	\includegraphics[width=0.5\linewidth]{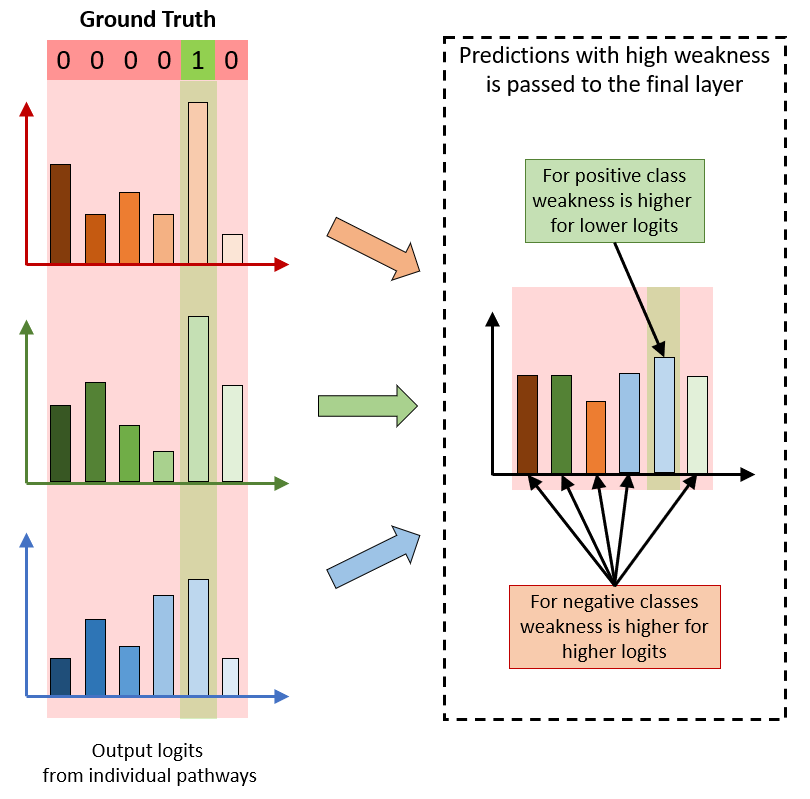}
	\caption{Generating combined logits by routing the weakest components corresponding to each class}
	\label{fig:prob}
\end{figure}
\begin{figure*}[htbp]
	\centering
	\includegraphics[width=\textwidth]{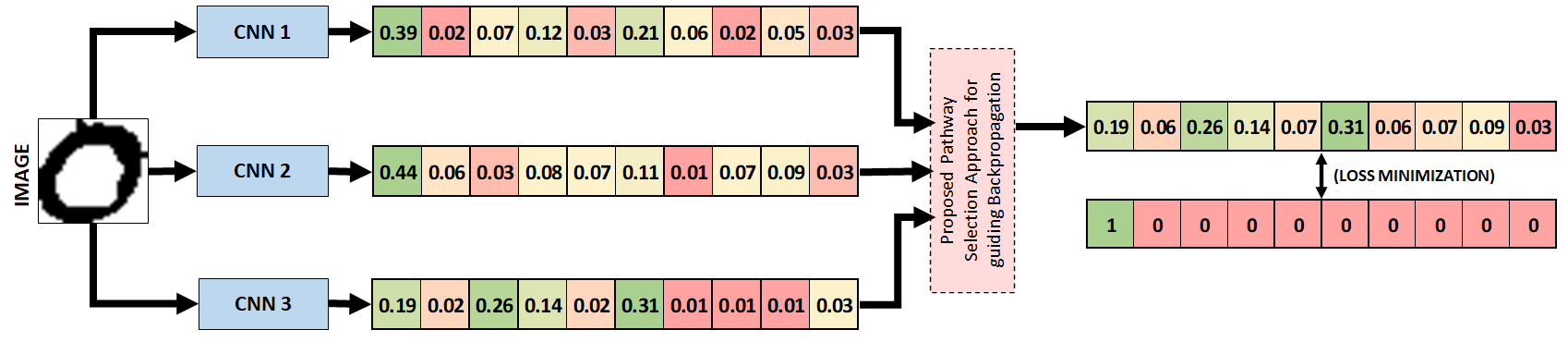}
	\caption{Model \textbf{M1}: Logits drawn from multiple instances of a model. ResNet18s\cite{resnet} were used in the experiments as the backend CNN}
	\label{f:ens}
\end{figure*}
\subsection{Inference}
\label{s:inf}
Once the multi-path network has been trained, the inference can be carried out using one of the two methods mentioned below.
\subsubsection{Combining strongest components of each pathway}
\label{s:inf1}

During evaluation of the network, the least weakest components are selected from each branch. Since the target vector $\mathbf{t}$ is not available during evaluation, a pseudo target vector $\hat{\mathbf{t}}$ is composed as,
\begin{equation}
\hat{\mathbf{t}}_i= \begin{cases}
1,  & \text{if } \hat{\mathbf{u}}_{ij} = \max_j(\hat{\mathbf{u}}_{ij})\\
0,  & \text{otherwise}
\end{cases}
\end{equation}
where, $i \in \{1,C\} \text{ and } j\in \{1,N\}$. The weakness function is now computed with respect to the pseudo target vector $\hat{\mathbf{t}}$ as

\begin{equation}
W(\hat{\mathbf{u}}_{ij}) = \hat{\mathbf{t}}_i+(-1)^{(1-\hat{\mathbf{t}}_i)}\times \frac{\hat{\mathbf{u}}_{ij}}{\sum_{m=1}^N{\hat{\mathbf{u}}_{im}}},
\end{equation}
where, $i \in \{1,C\} \text{ and } j\in \{1,N\}$. The final logits $\hat{\mathbf{v}}$ is calculated with respect to pathways with minimum weakness value as follows:
\begin{equation}
\hat{\mathbf{v}}_{i} = \argmin_{j}\  W({\hat{\mathbf{u}}_{ij}})
\end{equation}
where, $i \in \{1,C\} \text{ and } j\in \{1,N\}$.

\subsubsection{Averaging the logits from each pathway}
\label{s:inf2}
Multi-path networks can be trained by adding or averaging logits from each path and back propagating with equal preference. However, in the proposed method, the network focuses on improving the weakest components of each logit vector in every iteration. Hence, at the end of the training, the proposed method provides stronger individual branches as compared to the traditional method of adding or averaging logits. After training, if a mean of logits is considered for inference purpose a significant boost can be seen as the individual pathways have lesser weaknesses. In this way the resultant logits for inference may be written as
\begin{equation}
\hat{\mathbf{v}}_{i} = \frac{1}{N} \sum_j {\hat{\mathbf{u}}_{ij}}
\end{equation}
where, $i \in \{1,C\} \text{ and } j\in \{1,N\}$.

\subsection{Understanding the intuition} 

In standard column merging techniques like averaging or individual training, there is no constraint on the columns to recognize the strengths of the concepts learned by them. While averaging multiple columns often improve performance, it depends on the performance individual columns as well. It is often said that a team is as strong as its weakest player. Hence if one of the pathways is weaker than compared to the others, the effect of averaging would be diminished. The proposed methodology aims to address this. At any given iterations, the wrongest predictions would be attended to at first. This ensures all the classifier to perform at a similar level. This, in turn, improves the effect of column merging. 

\begin{figure}[h]
	\centering
	\includegraphics[width=0.5\linewidth]{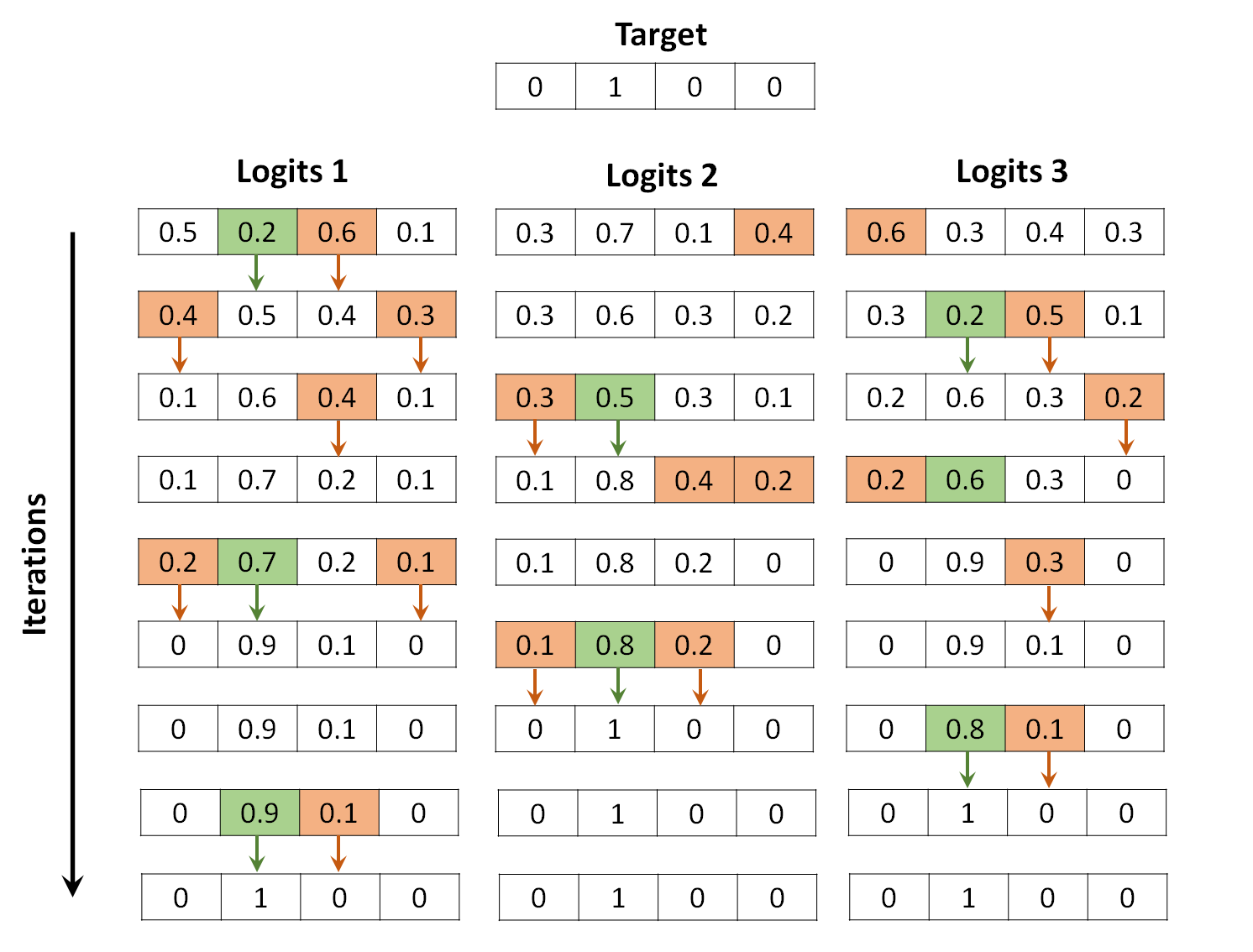}
	\caption{A simulation of logits progressing through iterations. Green arrows signify the positive class and red arrows signify negative classes.}
	\label{f:int}
\end{figure}

\par Fig. \ref{f:int} demonstrates a simulation of how the proposed approach can affect the logits in successive iterations. The values do not belong to an actual experiment but appropriately set for a better understanding of the intuition. The green arrows represent logits corresponding to the positive class. Here the value farthest from $1.0$ is attended at first while the others do not affect the gradients. The red arrows corresponding to the negative class attend to the values farthest from $0.0$.


\begin{figure*}[htbp]
	\centering
	\includegraphics[width=\textwidth]{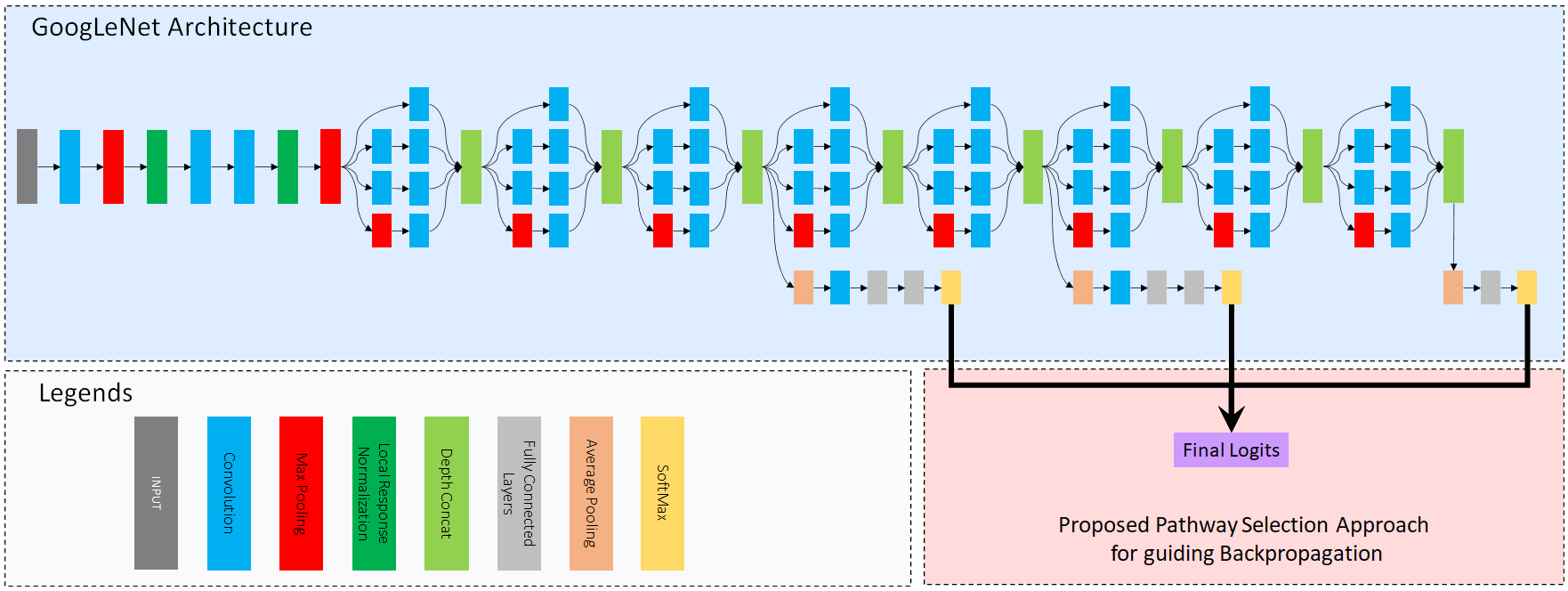}
	\caption{Model \textbf{M2}: Logits drawn from features with different scales. Based on a GoogLeNet\cite{googlenet} backend.}
	\label{fig:googlenet}
\end{figure*}

\begin{figure*}[htbp]
	\centering
	\includegraphics[width=\textwidth]{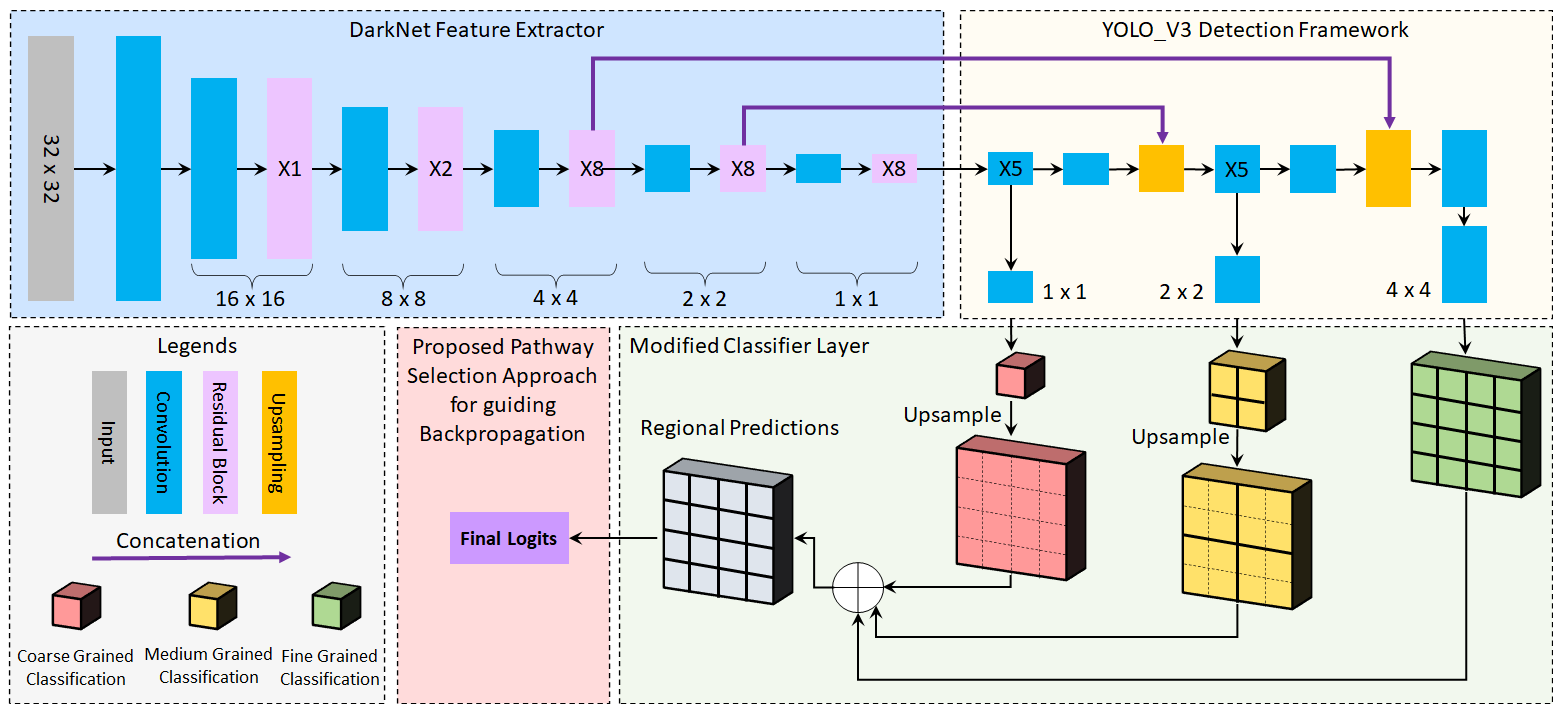}
	\caption{Model \textbf{M3}: Logits drawn from sub-regions in the feature space. Based on a DarkNet and YOLO\_v3 \cite{yolo} backend}
	\label{fig:yolo}
\end{figure*}

\begin{figure*}[h]
	\centering
	\includegraphics[width=\textwidth]{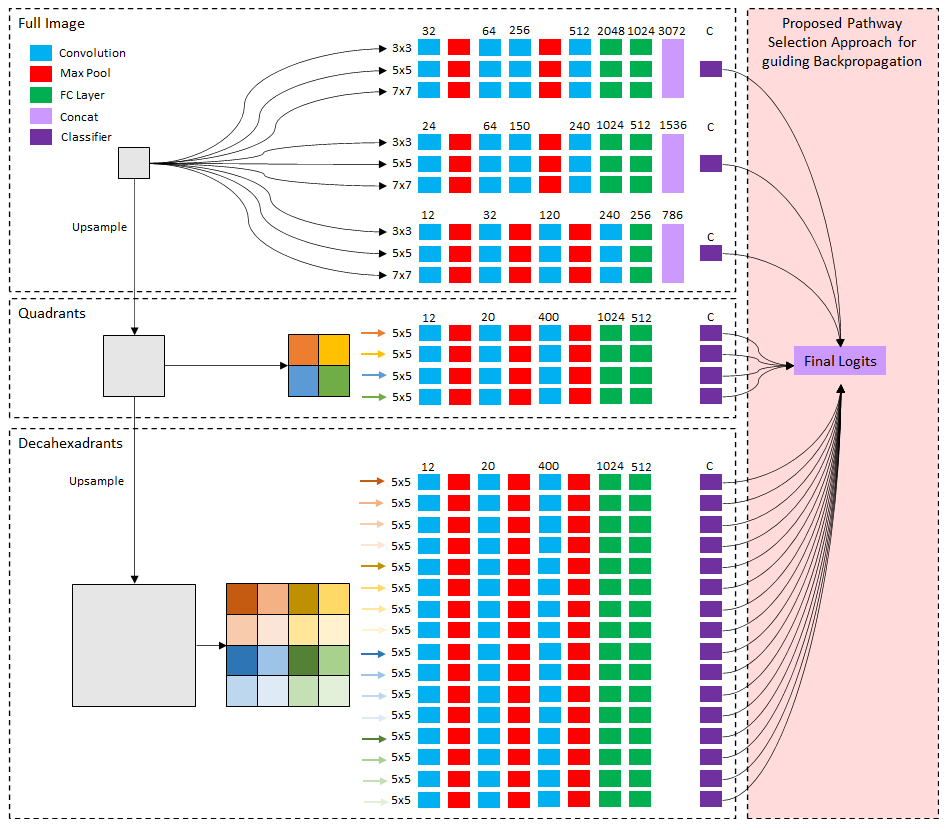}
	\caption{Model \textbf{M4}: Logits drawn from sub-regions in the input space. Inspired from the Quad Tree based architecture\cite{sarkhel2017multi}}
	\label{fig:quad}
\end{figure*}
\section{Applications}
\label{s:app}
The proposed methodology can be applied in multiple scenarios wherever we need to combine logits coming from different path. Till now the available options for dealing with logits was to either train them individually and use classifier ensemble techniques or compute the sum or mean of the logits to calculate the training loss. In the current approach, we have demonstrated four use cases where the proposed method can be used. The four use cases cover the various level of depths and locality of features. 

\subsection{Logits from multiple instances of a model (M1)}
\label{s:app1}

One of the most straight forward implementations of the proposed methodology is to use the method as an ensemble technique. Here we take log-softmax of the logits from individual CNN based classifier as input. The proposed approach provides us with a modified logits composed of the weakest components from each classifier. The backpropagation thus forces each of the CNNs to update its weight to enhance their weakest concepts. Fig. \ref{f:ens} demonstrates an example where outputs from three classifiers are combined using the proposed approach. Unlike traditional ensemble techniques, it is not necessary to train the networks individually. Simply averaging logits does not force the entire network to train explicitly with respect to the weaker components. In our experiments, we have combined logits from three 18 layer resnet models.

\subsection{Logits from different layers of a model (M2)}
\label{s:app2}

For very deep networks auxiliary classifiers often serve as gradient boosters for earlier layers. In the GoogLeNet model, two auxiliary classifiers are implemented during training in addition to the final classifier. The auxiliary classifiers are drawn from the end of the third and fifth inception blocks. However, these classifiers serve no purpose during the inference phase is hence chopped from the network once the training completes. These logits are drawn from layers with different levels of complexity of features. Obviously deeper networks will produce much smaller feature maps that model global concepts as compared to shallower features which capture granular features in larger feature maps. The proposed approach aims to bank on this factor to capture both shallow and deep features and train the preceding weights exclusively with respect to weaker components of the predicted logits. Using the proposed approach, the final logits and the two auxiliary logits are combined during training. Unlike the original network, the auxiliary classifiers contribute during the inference as well as mentioned in section \ref{s:inf}.

\renewcommand{\arraystretch}{1.2}
\begin{table*}[t]
	\centering
	\caption{Dataset Description}
	\label{t:dt}
	\resizebox{\textwidth}{!}{%
	\begin{tabular}{@{}llcccc@{}}
		\hline
		\textbf{Dataset Name}              & \textbf{Description}                                  & \textbf{Classes} & \textbf{Channels} & \textbf{Train Set} & \textbf{Test Set} \\ \hline \hline
		\textit{\textbf{CMATERdb   3.1.1}} & Bangla handwritten digits                             & 10               & 1                   & 4000               & 2000            \\
		\textit{\textbf{CMATERdb   3.2.1}} & Devanagari handwritten digits                         & 10               & 1                   & 2000               & 1000            \\
		\textit{\textbf{CMATERdb   3.3.1}} & Arabic handwritten digits                             & 10               & 1                   & 2000               & 1000            \\
		\textit{\textbf{CMATERdb   3.4.1}} & Telugu handwritten digits                             & 10               & 1                   & 2000               & 1000            \\
		\textit{\textbf{CMATERdb   3.1.2}} & Bangla handwritten basic-characters                   & 50               & 1                   & 12000              & 3000            \\
		\textit{\textbf{CMATERdb   3.1.3}} & Bangla handwritten compound-characters                & 199              & 1                   & 34229              & 8468            \\
		\textit{\textbf{MNIST}}            & English handwritten digits                            & 10               & 1                   & 60000              & 10000           \\
		\textit{\textbf{Fashion-MNIST}}    & MNIST like dataset of fashion products                & 10               & 1                   & 60000              & 10000           \\
		\textit{\textbf{notMNIST}}         & MNIST like dataset of A to J in various typefaces     & 10               & 1                   & 60000              & 10000           \\
		\textit{\textbf{Kannada-MNIST}}     & MNIST like dataset of Kannada handwritten digits      & 10               & 1                   & 60000              & 10000           \\
		\textit{\textbf{Kuzushiji-MNIST}}  & MNIST like dataset of Japanese handwritten characters & 10               & 1                   & 60000              & 10000           \\
		\textit{\textbf{SVHN}}             & English digit dataset from street view images         & 10               & 3                   & 73257              & 26032           \\
		\textit{\textbf{Cifar10}}          & Images of objects occurring in natural scenes         & 10               & 3                   & 50000              & 10000           \\
		\hline
	\end{tabular}%
		}
\end{table*}

\subsection{Logits from feature subspace (M3)}
\label{s:app3}

Objects can appear in different sizes and position throughout the space of the image and this makes it particularly hard for object detection approaches to locate the object in the scene. This challenge also exists for object recognition tasks. Different size and position of objects can be activated not only features at different depths but also at different positions. Networks like YOLO\_V3 divides features of various depths into grids of different sizes which are then used to locate centres of bounding boxes. Our proposition modifies the classification layer of the network. With an input of $32\times32$ the network produces three regional predictions of size $1\times1$, $2\times2$, and $4\times4$, which correspond to the entire feature map, each quadrant and each decahexadrants of the feature map respectively. Buy up-scaling the $1\times1$ and $2\times2$ feature maps to $4\times4$ using nearest neighbor interpolation and adding them we can obtain 16 predictions corresponding to 16 different regions of the feature space. The proposed approach can then combine the 16 prediction vectors corresponding to the $4\times4$ grid of the feature space to form the final logits. In this case, the model attempts to train different part of the networks to attend to the weakness of different regional concepts with respect to the different classes.

\subsection{Logits from input subspace(M4)}
\label{s:app4}

Just like the YOLO based network computed the logits from sub-regions in the feature space, it is also possible to divide the inputs to different regions and run parallel multi column deep neural networks on individual regions. The columns can either be combined by averaging the softmax of the logits or by averaging them. In both cases the individual networks set no preference on the strength of the concepts learned by them. The network proposed in the works of [CITE] combines 23 CNNs working on regions of different sizes corresponding to different positions. The proposed approach combines those logits to create a model where all columns can be trained together and they focus on strengthening their weakest concepts.


\renewcommand{\arraystretch}{1.5}
\begin{table*}[h]
	\centering
	\caption{Results of all experiments}
	\label{t:res}
    \resizebox{\textwidth}{!}{%
		\begin{tabular}{lccccccccccccccccc}
			\Xhline{2pt}
			& \multicolumn{4}{c}{\textbf{Logits from multiple instances of a model (M1)}}                                                                                                                                                                                                                & \textbf{} & \multicolumn{4}{c}{\textbf{Logits from different layers of a model (M2)}}                                                                                                                                                                                                                   & \textbf{}                        & \multicolumn{3}{c}{\textbf{Logits from feature subspaces (M3)}}                                                                                                                                                       & \textbf{}                        & \multicolumn{3}{c}{\textbf{Logits from input subspaces (M4)}}                                                                                                                                                            \\ \cline{2-18} 
			\multirow{-2}{*}{\textbf{Datasets}}  & \textbf{\begin{tabular}[c]{@{}c@{}}Single\\ ResNet18\end{tabular}} & \textbf{\begin{tabular}[c]{@{}c@{}}Average of\\ 3 Logits\end{tabular}} & \textbf{\begin{tabular}[c]{@{}c@{}}Strong\\ Inference\end{tabular}} & \textbf{\begin{tabular}[c]{@{}c@{}}Mean\\ Inference\end{tabular}} & \textbf{} & \textbf{\begin{tabular}[c]{@{}c@{}}Native\\ GoogLeNet\end{tabular}} & \textbf{\begin{tabular}[c]{@{}c@{}}Average of\\ 3 Logits\end{tabular}} & \textbf{\begin{tabular}[c]{@{}c@{}}Strong\\ Inference\end{tabular}} & \textbf{\begin{tabular}[c]{@{}c@{}}Mean\\ Inference\end{tabular}} & \textbf{}                        & \textbf{\begin{tabular}[c]{@{}c@{}}Average of\\ 16 grids\end{tabular}} & \textbf{\begin{tabular}[c]{@{}c@{}}Strong\\ Inference\end{tabular}} & \textbf{\begin{tabular}[c]{@{}c@{}}Mean\\ Inference\end{tabular}} & \textbf{}                        & \textbf{\begin{tabular}[c]{@{}c@{}}Average of\\ 23 Branches\end{tabular}} & \textbf{\begin{tabular}[c]{@{}c@{}}Strong\\ Inference\end{tabular}} & \textbf{\begin{tabular}[c]{@{}c@{}}Mean\\ Inference\end{tabular}} \\ \hline \hline
			\textit{\textbf{CMATERdb 3.1.1}}             & \cellcolor[HTML]{FA9C9E}{\color[HTML]{000000} 99.20}               & \cellcolor[HTML]{FDDEB8}{\color[HTML]{000000} 99.50}                   & \cellcolor[HTML]{D6E9BE}{\color[HTML]{000000} 99.70}                & \cellcolor[HTML]{ADDDBA}{\color[HTML]{000000} \textbf{99.80}}     & \textbf{} & \cellcolor[HTML]{FA9C9E}{\color[HTML]{000000} 96.70}                & \cellcolor[HTML]{FEEEBE}{\color[HTML]{000000} 98.70}                   & \cellcolor[HTML]{E4EDBF}{\color[HTML]{000000} 99.00}                & \cellcolor[HTML]{ADDDBA}{\color[HTML]{000000} 99.30}              & {\color[HTML]{000000} }          & \cellcolor[HTML]{FA9C9E}{\color[HTML]{000000} 98.70}                   & \cellcolor[HTML]{FFF5C1}{\color[HTML]{000000} 99.70}                & \cellcolor[HTML]{ADDDBA}{\color[HTML]{000000} \textbf{99.80}}     & {\color[HTML]{000000} \textbf{}} & \cellcolor[HTML]{FA9C9E}{\color[HTML]{000000} 99.10}                      & \cellcolor[HTML]{FFF5C1}{\color[HTML]{000000} 99.40}                & \cellcolor[HTML]{ADDDBA}{\color[HTML]{000000} 99.50}              \\
			\textit{\textbf{CMATERdb 3.2.1}}            & \cellcolor[HTML]{FA9C9E}{\color[HTML]{000000} 99.30}               & \cellcolor[HTML]{FEF0BF}{\color[HTML]{000000} 99.48}                   & \cellcolor[HTML]{ADDDBA}{\color[HTML]{000000} 99.55}                & \cellcolor[HTML]{F2F2C0}{\color[HTML]{000000} 99.50}              &           & \cellcolor[HTML]{FA9C9E}{\color[HTML]{000000} 97.50}                & \cellcolor[HTML]{FFF5C1}{\color[HTML]{000000} 99.35}                   & \cellcolor[HTML]{FFF5C1}{\color[HTML]{000000} 99.35}                & \cellcolor[HTML]{ADDDBA}{\color[HTML]{000000} 99.50}              & {\color[HTML]{000000} }          & \cellcolor[HTML]{FA9C9E}{\color[HTML]{000000} 99.40}                   & \cellcolor[HTML]{FFF5C1}{\color[HTML]{000000} 99.60}                & \cellcolor[HTML]{ADDDBA}{\color[HTML]{000000} \textbf{99.85}}     & {\color[HTML]{000000} \textbf{}} & \cellcolor[HTML]{FA9C9E}{\color[HTML]{000000} 99.10}                      & \cellcolor[HTML]{FFF5C1}{\color[HTML]{000000} 99.30}                & \cellcolor[HTML]{ADDDBA}{\color[HTML]{000000} 99.60}              \\
			\textit{\textbf{CMATERdb 3.3.1}}              & \cellcolor[HTML]{FA9C9E}{\color[HTML]{000000} 99.40}               & \cellcolor[HTML]{FCC1AC}{\color[HTML]{000000} 99.48}                   & \cellcolor[HTML]{ADDDBA}{\color[HTML]{000000} 99.70}                & \cellcolor[HTML]{ADDDBA}{\color[HTML]{000000} 99.70}              &           & \cellcolor[HTML]{FA9C9E}{\color[HTML]{000000} 98.40}                & \cellcolor[HTML]{FDDBB7}{\color[HTML]{000000} 98.90}                   & \cellcolor[HTML]{DFECBF}{\color[HTML]{000000} 99.30}                & \cellcolor[HTML]{ADDDBA}{\color[HTML]{000000} 99.60}              & {\color[HTML]{000000} }          & \cellcolor[HTML]{FA9C9E}{\color[HTML]{000000} 99.30}                   & \cellcolor[HTML]{FFF5C1}{\color[HTML]{000000} 99.70}                & \cellcolor[HTML]{ADDDBA}{\color[HTML]{000000} \textbf{99.90}}     & {\color[HTML]{000000} \textbf{}} & \cellcolor[HTML]{FA9C9E}{\color[HTML]{000000} 98.70}                      & \cellcolor[HTML]{FFF5C1}{\color[HTML]{000000} 99.10}                & \cellcolor[HTML]{ADDDBA}{\color[HTML]{000000} 99.40}              \\
			\textit{\textbf{CMATERdb 3.4.1}}             & \cellcolor[HTML]{FA9C9E}{\color[HTML]{000000} 98.90}               & \cellcolor[HTML]{FDD7B5}{\color[HTML]{000000} 99.10}                   & \cellcolor[HTML]{E4EDBF}{\color[HTML]{000000} 99.30}                & \cellcolor[HTML]{ADDDBA}{\color[HTML]{000000} 99.50}              &           & \cellcolor[HTML]{FA9C9E}{\color[HTML]{000000} 97.20}                & \cellcolor[HTML]{FEE8BC}{\color[HTML]{000000} 98.70}                   & \cellcolor[HTML]{E0ECBF}{\color[HTML]{000000} 99.20}                & \cellcolor[HTML]{ADDDBA}{\color[HTML]{000000} 99.60}              & {\color[HTML]{000000} }          & \cellcolor[HTML]{FA9C9E}{\color[HTML]{000000} 99.00}                   & \cellcolor[HTML]{ADDDBA}{\color[HTML]{000000} \textbf{99.80}}       & \cellcolor[HTML]{ADDDBA}{\color[HTML]{000000} \textbf{99.80}}     & {\color[HTML]{000000} \textbf{}} & \cellcolor[HTML]{FA9C9E}{\color[HTML]{000000} 98.90}                      & \cellcolor[HTML]{FFF5C1}{\color[HTML]{000000} 99.20}                & \cellcolor[HTML]{ADDDBA}{\color[HTML]{000000} 99.70}              \\
			\textit{\textbf{CMATERdb 3.1.2}}      & \cellcolor[HTML]{FA9C9E}{\color[HTML]{000000} 97.67}               & \cellcolor[HTML]{FBB6A8}{\color[HTML]{000000} 97.80}                   & \cellcolor[HTML]{ADDDBA}{\color[HTML]{000000} 98.40}                & \cellcolor[HTML]{ADDDBA}{\color[HTML]{000000} 98.40}              &           & \cellcolor[HTML]{FA9C9E}{\color[HTML]{000000} 98.00}                & \cellcolor[HTML]{FFF5C1}{\color[HTML]{000000} 98.23}                   & \cellcolor[HTML]{FFF5C1}{\color[HTML]{000000} 98.23}                & \cellcolor[HTML]{ADDDBA}{\color[HTML]{000000} 98.33}              & {\color[HTML]{000000} }          & \cellcolor[HTML]{FA9C9E}{\color[HTML]{000000} 98.50}                   & \cellcolor[HTML]{FFF5C1}{\color[HTML]{000000} 98.63}                & \cellcolor[HTML]{ADDDBA}{\color[HTML]{000000} \textbf{98.90}}     & {\color[HTML]{000000} \textbf{}} & \cellcolor[HTML]{FA9C9E}{\color[HTML]{000000} 97.67}                      & \cellcolor[HTML]{FFF5C1}{\color[HTML]{000000} 97.80}                & \cellcolor[HTML]{ADDDBA}{\color[HTML]{000000} 98.00}              \\
			\textit{\textbf{CMATERdb 3.1.3}} & \cellcolor[HTML]{FA9C9E}{\color[HTML]{000000} 96.40}               & \cellcolor[HTML]{FBBDAB}{\color[HTML]{000000} 96.70}                   & \cellcolor[HTML]{ADDDBA}{\color[HTML]{000000} \textbf{97.90}}       & \cellcolor[HTML]{C5E4BD}{\color[HTML]{000000} 97.70}              &           & \cellcolor[HTML]{FA9C9E}{\color[HTML]{000000} 96.90}                & \cellcolor[HTML]{FEF3C0}{\color[HTML]{000000} 97.32}                   & \cellcolor[HTML]{FEF5C1}{\color[HTML]{000000} 97.33}                & \cellcolor[HTML]{ADDDBA}{\color[HTML]{000000} 97.67}              & {\color[HTML]{000000} }          & \cellcolor[HTML]{FA9C9E}{\color[HTML]{000000} 97.20}                   & \cellcolor[HTML]{ADDDBA}{\color[HTML]{000000} \textbf{97.90}}       & \cellcolor[HTML]{FFF5C1}{\color[HTML]{000000} 97.70}              & {\color[HTML]{000000} }          & \cellcolor[HTML]{FA9C9E}{\color[HTML]{000000} 94.70}                      & \cellcolor[HTML]{FFF5C1}{\color[HTML]{000000} 96.32}                & \cellcolor[HTML]{ADDDBA}{\color[HTML]{000000} 96.50}              \\
			\textit{\textbf{MNIST}}              & \cellcolor[HTML]{FA9C9E}{\color[HTML]{000000} 99.63}               & \cellcolor[HTML]{FA9C9E}{\color[HTML]{000000} 99.63}                   & \cellcolor[HTML]{D3E9BE}{\color[HTML]{000000} 99.70}                & \cellcolor[HTML]{ADDDBA}{\color[HTML]{000000} \textbf{99.73}}     & \textbf{} & \cellcolor[HTML]{FA9C9E}{\color[HTML]{000000} 99.58}                & \cellcolor[HTML]{FAA7A2}{\color[HTML]{000000} 99.59}                   & \cellcolor[HTML]{ADDDBA}{\color[HTML]{000000} 99.73}                & \cellcolor[HTML]{ADDDBA}{\color[HTML]{000000} \textbf{99.73}}     & {\color[HTML]{000000} \textbf{}} & \cellcolor[HTML]{FA9C9E}{\color[HTML]{000000} 99.55}                   & \cellcolor[HTML]{ADDDBA}{\color[HTML]{000000} 99.65}                & \cellcolor[HTML]{ADDDBA}{\color[HTML]{000000} 99.65}              & {\color[HTML]{000000} }          & \cellcolor[HTML]{FFF5C1}{\color[HTML]{000000} 99.29}                      & \cellcolor[HTML]{FA9C9E}{\color[HTML]{000000} 99.25}                & \cellcolor[HTML]{ADDDBA}{\color[HTML]{000000} 99.40}              \\
			\textit{\textbf{Fashion MNIST}}      & \cellcolor[HTML]{FA9C9E}{\color[HTML]{000000} 94.00}               & \cellcolor[HTML]{FEE5BA}{\color[HTML]{000000} 94.28}                   & \cellcolor[HTML]{E1ECBF}{\color[HTML]{000000} 94.40}                & \cellcolor[HTML]{ADDDBA}{\color[HTML]{000000} 94.50}              &           & \cellcolor[HTML]{FA9C9E}{\color[HTML]{000000} 94.30}                & \cellcolor[HTML]{FEE8BC}{\color[HTML]{000000} 94.45}                   & \cellcolor[HTML]{ADDDBA}{\color[HTML]{000000} 94.53}                & \cellcolor[HTML]{DAEBBE}{\color[HTML]{000000} 94.50}              & {\color[HTML]{000000} }          & \cellcolor[HTML]{FA9C9E}{\color[HTML]{000000} 92.26}                   & \cellcolor[HTML]{FFF5C1}{\color[HTML]{000000} 94.01}                & \cellcolor[HTML]{ADDDBA}{\color[HTML]{000000} \textbf{94.60}}     & {\color[HTML]{000000} }          & \cellcolor[HTML]{FFF5C1}{\color[HTML]{000000} 93.09}                      & \cellcolor[HTML]{FA9C9E}{\color[HTML]{000000} 92.40}                & \cellcolor[HTML]{ADDDBA}{\color[HTML]{000000} 94.00}              \\
			\textit{\textbf{notMNIST}}           & \cellcolor[HTML]{FA9C9E}{\color[HTML]{000000} 96.80}               & \cellcolor[HTML]{FBBCAA}{\color[HTML]{000000} 97.00}                   & \cellcolor[HTML]{B2DFBB}{\color[HTML]{000000} 97.71}                & \cellcolor[HTML]{ADDDBA}{\color[HTML]{000000} \textbf{97.73}}     &           & \cellcolor[HTML]{FA9C9E}{\color[HTML]{000000} 97.12}                & \cellcolor[HTML]{FDD8B5}{\color[HTML]{000000} 97.42}                   & \cellcolor[HTML]{B7E0BB}{\color[HTML]{000000} 97.71}                & \cellcolor[HTML]{ADDDBA}{\color[HTML]{000000} \textbf{97.73}}     & {\color[HTML]{000000} }          & \cellcolor[HTML]{FA9C9E}{\color[HTML]{000000} 97.02}                   & \cellcolor[HTML]{FFF5C1}{\color[HTML]{000000} 97.09}                & \cellcolor[HTML]{ADDDBA}{\color[HTML]{000000} 97.33}              & {\color[HTML]{000000} }          & \cellcolor[HTML]{FA9C9E}{\color[HTML]{000000} 95.98}                      & \cellcolor[HTML]{FFF5C1}{\color[HTML]{000000} 96.02}                & \cellcolor[HTML]{ADDDBA}{\color[HTML]{000000} 96.67}              \\
			\textit{\textbf{Kannada-MNIST}}      & \cellcolor[HTML]{FA9C9E}{\color[HTML]{000000} 97.40}               & \cellcolor[HTML]{FBBDAB}{\color[HTML]{000000} 97.63}                   & \cellcolor[HTML]{ADDDBA}{\color[HTML]{000000} \textbf{98.87}}       & \cellcolor[HTML]{DBEBBE}{\color[HTML]{000000} 98.40}              &           & \cellcolor[HTML]{FA9C9E}{\color[HTML]{000000} 96.67}                & \cellcolor[HTML]{FBB7A8}{\color[HTML]{000000} 97.02}                   & \cellcolor[HTML]{ADDDBA}{\color[HTML]{000000} \textbf{98.87}}       & \cellcolor[HTML]{C2E4BC}{\color[HTML]{000000} 98.60}              & {\color[HTML]{000000} }          & \cellcolor[HTML]{FA9C9E}{\color[HTML]{000000} 96.67}                   & \cellcolor[HTML]{ADDDBA}{\color[HTML]{000000} 98.85}                & \cellcolor[HTML]{FFF5C1}{\color[HTML]{000000} 98.50}              & {\color[HTML]{000000} }          & \cellcolor[HTML]{FA9C9E}{\color[HTML]{000000} 96.40}                      & \cellcolor[HTML]{ADDDBA}{\color[HTML]{000000} 98.54}                & \cellcolor[HTML]{FFF5C1}{\color[HTML]{000000} 98.50}              \\
			\textit{\textbf{Kuzushiji-MNIST}}    & \cellcolor[HTML]{FA9C9E}{\color[HTML]{000000} 96.67}               & \cellcolor[HTML]{FDD2B3}{\color[HTML]{000000} 97.20}                   & \cellcolor[HTML]{C4E4BC}{\color[HTML]{000000} 97.87}                & \cellcolor[HTML]{ADDDBA}{\color[HTML]{000000} 98.00}              &           & \cellcolor[HTML]{FA9C9E}{\color[HTML]{000000} 97.40}                & \cellcolor[HTML]{FCCCB0}{\color[HTML]{000000} 97.66}                   & \cellcolor[HTML]{ADDDBA}{\color[HTML]{000000} \textbf{98.10}}       & \cellcolor[HTML]{ADDDBA}{\color[HTML]{000000} \textbf{98.10}}     & {\color[HTML]{000000} }          & \cellcolor[HTML]{FA9C9E}{\color[HTML]{000000} 97.20}                   & \cellcolor[HTML]{FFF5C1}{\color[HTML]{000000} 97.61}                & \cellcolor[HTML]{ADDDBA}{\color[HTML]{000000} 97.67}              & {\color[HTML]{000000} }          & \cellcolor[HTML]{FA9C9E}{\color[HTML]{000000} 96.00}                      & \cellcolor[HTML]{FFF5C1}{\color[HTML]{000000} 96.77}                & \cellcolor[HTML]{ADDDBA}{\color[HTML]{000000} 97.20}              \\
			\textit{\textbf{Svhn}}               & \cellcolor[HTML]{FA9C9E}{\color[HTML]{000000} 95.50}               & \cellcolor[HTML]{FCC9AF}{\color[HTML]{000000} 95.67}                   & \cellcolor[HTML]{ADDDBA}{\color[HTML]{000000} 96.00}                & \cellcolor[HTML]{ADDDBA}{\color[HTML]{000000} 96.00}              &           & \cellcolor[HTML]{FA9C9E}{\color[HTML]{000000} 95.82}                & \cellcolor[HTML]{FAACA4}{\color[HTML]{000000} 95.85}                   & \cellcolor[HTML]{CCE6BD}{\color[HTML]{000000} 96.12}                & \cellcolor[HTML]{ADDDBA}{\color[HTML]{000000} \textbf{96.20}}     & {\color[HTML]{000000} }          & \cellcolor[HTML]{FA9C9E}{\color[HTML]{000000} 84.19}                   & \cellcolor[HTML]{FFF5C1}{\color[HTML]{000000} 85.94}                & \cellcolor[HTML]{ADDDBA}{\color[HTML]{000000} 86.40}              & {\color[HTML]{000000} }          & \cellcolor[HTML]{FA9C9E}{\color[HTML]{000000} 89.47}                      & \cellcolor[HTML]{FFF5C1}{\color[HTML]{000000} 89.86}                & \cellcolor[HTML]{ADDDBA}{\color[HTML]{000000} 91.33}              \\
			\textit{\textbf{Cifar10}}            & \cellcolor[HTML]{FA9C9E}{\color[HTML]{000000} 94.19}               & \cellcolor[HTML]{FDE1B9}{\color[HTML]{000000} 94.59}                   & \cellcolor[HTML]{ECF0C0}{\color[HTML]{000000} 94.82}                & \cellcolor[HTML]{ADDDBA}{\color[HTML]{000000} \textbf{95.20}}     &                         & \cellcolor[HTML]{FA9C9E}{\color[HTML]{000000} 91.94}                & \cellcolor[HTML]{FBB8A9}{\color[HTML]{000000} 92.02}                   & \cellcolor[HTML]{D3E8BE}{\color[HTML]{000000} 92.36}                & \cellcolor[HTML]{ADDDBA}{\color[HTML]{000000} 92.50}              & {\color[HTML]{000000} }          & \cellcolor[HTML]{FA9C9E}{\color[HTML]{000000} 72.80}                   & \cellcolor[HTML]{FFF5C1}{\color[HTML]{000000} 73.77}                & \cellcolor[HTML]{ADDDBA}{\color[HTML]{000000} 75.00}              & {\color[HTML]{000000} }          & \cellcolor[HTML]{FA9C9E}{\color[HTML]{000000} 84.53}                      & \cellcolor[HTML]{FFF5C1}{\color[HTML]{000000} 84.67}                & \cellcolor[HTML]{ADDDBA}{\color[HTML]{000000} 85.00}              \\ \hline
			\textit{\textbf{Mean}}               & \cellcolor[HTML]{E8EEBF}{\color[HTML]{000000} 97.31}               & \cellcolor[HTML]{D4E9BE}{\color[HTML]{000000} 97.54}                   & \cellcolor[HTML]{AFDEBB}{\color[HTML]{000000} 97.99}                & \cellcolor[HTML]{ADDDBA}{\color[HTML]{000000} \textbf{98.01}}              & {\color[HTML]{000000} } & \cellcolor[HTML]{FA9C9E}{\color[HTML]{000000} 96.73}                & \cellcolor[HTML]{FDE0B8}{\color[HTML]{000000} 97.32}                   & \cellcolor[HTML]{CEE7BD}{\color[HTML]{000000} 97.68}                & \cellcolor[HTML]{ADDDBA}{\color[HTML]{000000} \textbf{97.80}}              & {\color[HTML]{000000} }          & \cellcolor[HTML]{FA9C9E}{\color[HTML]{000000} 94.75}                   & \cellcolor[HTML]{FFF5C1}{\color[HTML]{000000} 95.56}                & \cellcolor[HTML]{ADDDBA}{\color[HTML]{000000} \textbf{95.78}}              & {\color[HTML]{000000} }          & \cellcolor[HTML]{FA9C9E}{\color[HTML]{000000} 95.61}                      & \cellcolor[HTML]{FFF5C1}{\color[HTML]{000000} 96.05}                & \cellcolor[HTML]{ADDDBA}{\color[HTML]{000000} \textbf{96.52}}              \\ \Xhline{2pt}
		\end{tabular}%
	}
\end{table*}
\section{Results and Discussions}
Several experiments were carried out to validate the method of combination of multi column networks. The focus of the experiments was to prove that by training multi column models by the proposed method is more efficient than by averaging the outputs of the different columns. The inference was carried out by two different ways as mentioned section \ref{s:inf1} and \ref{s:inf2}. The proposed method was tested in 4 different application scenarios as mentioned in section \ref{s:app1} - \ref{s:app4}. The experiments were carried out on 14 different datasets that can be categorized based on various parameters.

\subsection{Dataset Descriptions}
The CMATERdb datasets consist of handwritten digits and characters from Indic scripts. There are 4 handwritten digit datasets with around a 200-400 hundred samples per class. Each of the digit datasets has samples divided into 10 classes corresponding to the digits. Additionally, there are 2 datasets focusing on characters. One of them has 50 classes and 240 samples per class and the other has 199 classes and around 172 samples per class. These datasets with higher number of classes and comparatively lower number of samples pose a serious challenge. We also look in datasets with the higher number of samples as well. Along with the MNIST dataset, there are 4 other datasets that follow the distribution pattern of MNIST. Each of them consists of 10 classes and around 6000 samples per class and focus on handwritten digits and characters from Kannada and Japanese, English alphabets in various typefaces and images related to fashion. All the previously mentioned datasets were in grayscale. Three RGB datasets were also considered. The SVHN dataset consists of digits occurring in natural scenes. It has 10 classes and  average of 7325.7 training samples per class. Finally, the CIFAR 10 dataset was considered where natural scene objects are divided into 10 categories. No dataset augmentations were used but the inputs were normalized to a zero mean and unity variance. The dataset descriptions can be found in table \ref{t:dt}.
	\begin{figure}[h]
		\centering
		\includegraphics[width=0.5\linewidth]{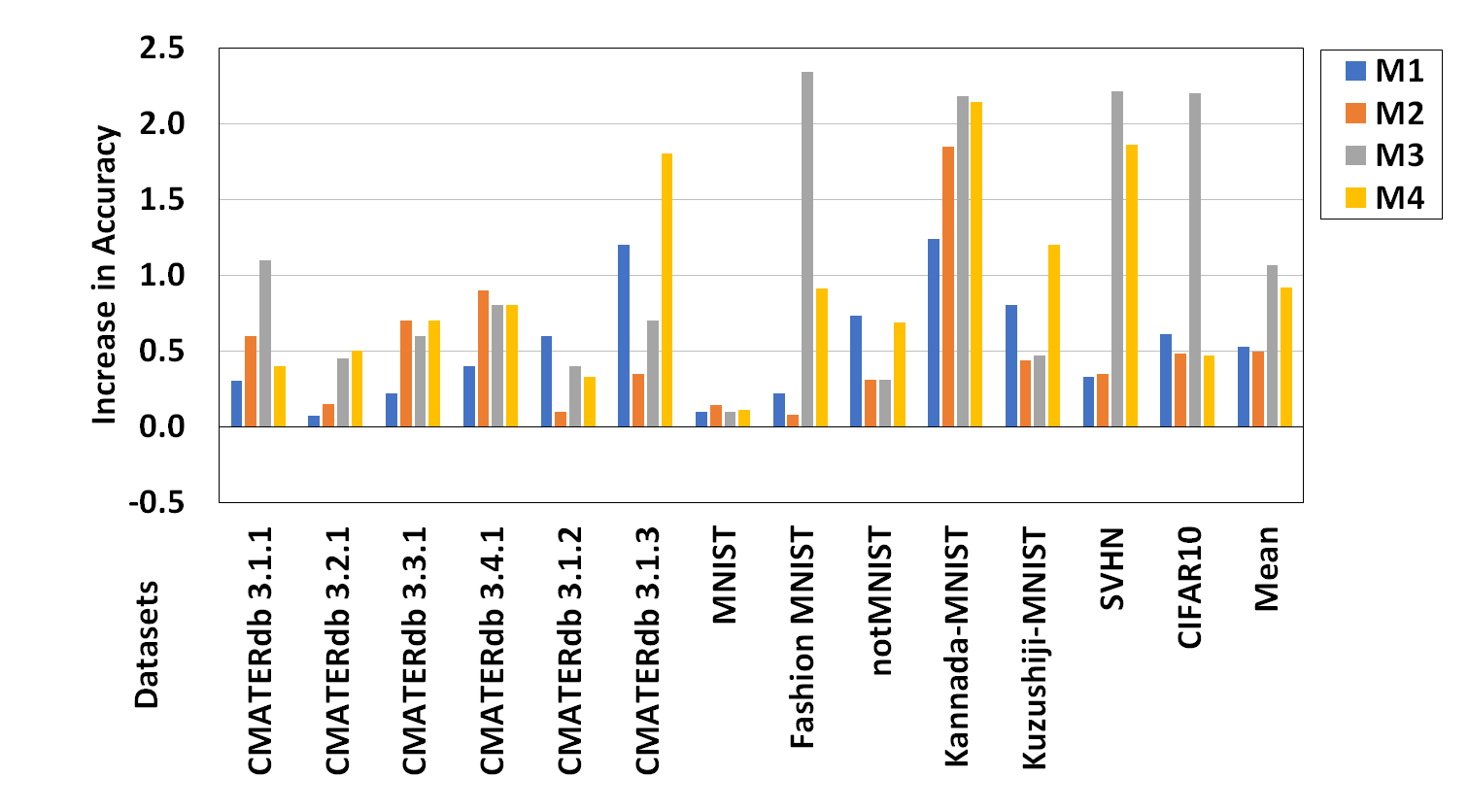}
		\includegraphics[width=0.5\linewidth]{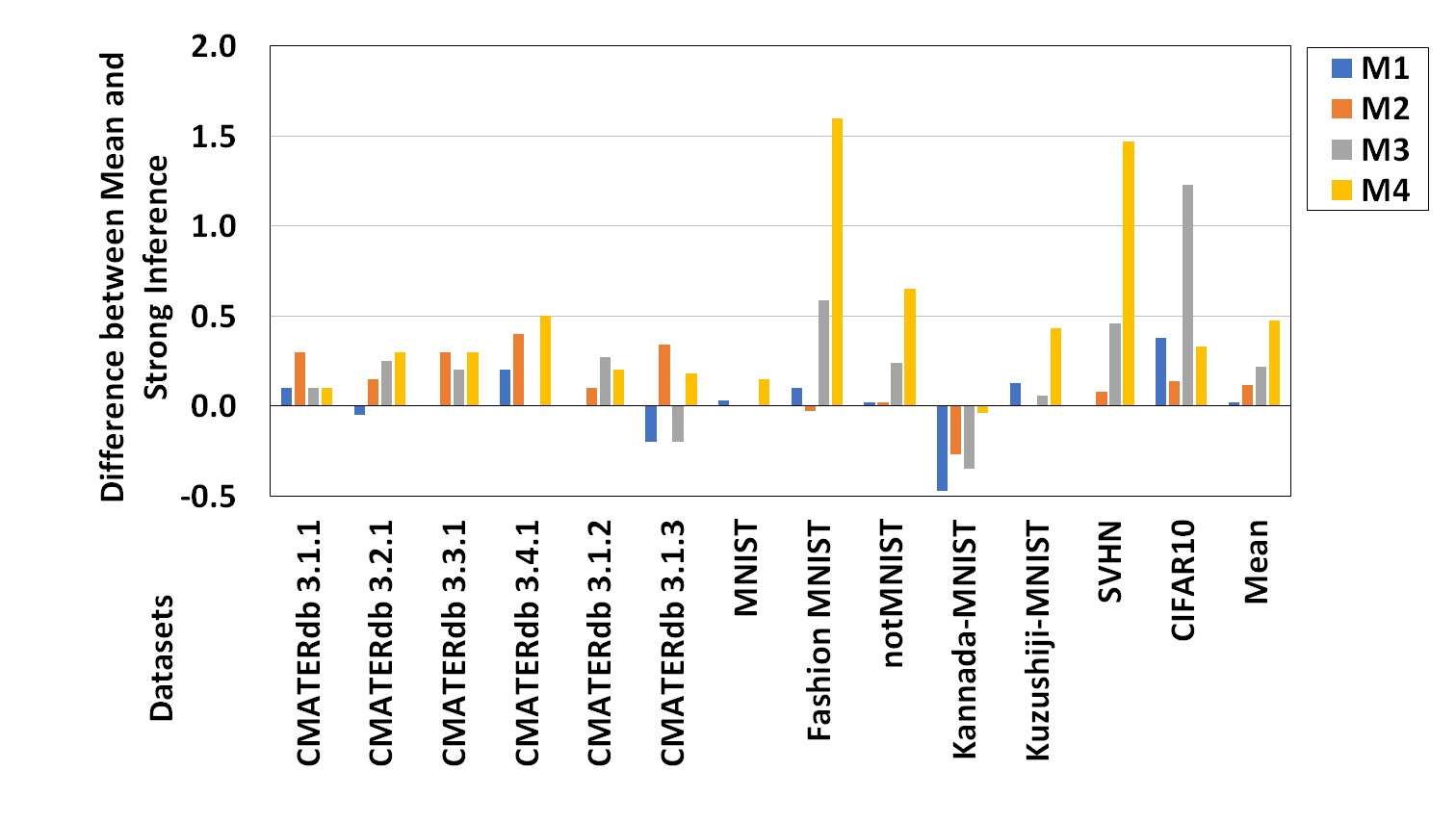}
		\caption{Increase in Accuracy of the best of mean and strong inference protocol with respect to traditional averaging of logits(Left), Difference between the Mean Inference Protocol vs the Strong Inference Protocol (Mean-Strong)(Right)}
		\label{f:inc}
	\end{figure}

\subsection{Experimental Results}
As mentioned before 4 models have been proposed in section \ref{s:app}. The 4 models are built upon partially or fully adapted versions resnet18, googlenet, YOLO\_v3 and the Quad Tree based MCDNN. No pre-trained models were used and all training was done from freshly initialized network. The networks were trained and the model with the best training accuracy was chosen for testing. The use of multiple columns itself has a regularization effect on the training process hence early stopping was not necessary. Adabound optimizer was used to update the weights to reduce the cross entropy loss between the combined logits and the ground truth. In each of the four proposed models, we have two inference protocols as mentioned in section \ref{s:inf}. "Strong Inference" corresponds to the method described in section \ref{s:inf1} and "Mean Inference" corresponds to the method described in section \ref{s:inf2}. In all the 4 cases, we have compared against a version of the models where we have simply averaged the logits using in the traditional way during training. Additionally, a comparison has also been against the native model as it was proposed.  The result of all the experiments is compiled in table \ref{t:res}. 

\subsection{Observations}
If we observe the results in the table \ref{t:res} we can draw some inferences with respect to the performance of the proposed models. The inferences are enumerated as follows:
\begin{enumerate}
	
	\item Either of the proposed inference protocols always triumph over the traditional method of averaging multiple logits. The observation is demonstrated in fig. \ref{f:inc}. The trend is also or native designs of the back-end network like standalone ResNet18 or GoogLeNet.

	\item The ''mean inference'' protocol performs best in several datasets, but the "strong inference" protocol has also shown better performance in some of them as well. This is shown in fig. \ref{fig:inc}.
	\item If the average trend is followed in table \ref{t:res}, M1 produces the best results followed by M2, M4 and M3 successively. Several trends can also be noticed under careful observations.
	\begin{enumerate}
		\item M3 produces the best performance for all the 6 CMATERdb datasets with lower number of samples. This trend is also seen for one of the larger gray-scale dataset namely Fashion-MNIST. Small datasets that adhere to a basic template, benefits from the region specific learning technique. The effect of feature subspace analysis is also particularly beneficial for Fashion-MNIST where the samples correspond fashion attires that generally follow a relatively strong predefined template with low variance.
		\item M2 produces the best results for 4 out of 5 MNIST like datasets and SVHN which is a RGB dataset. The intra-class variance for larger dataset is much higher hence feature subspace cannot be exploited as effectively. Fetching features from different layers can combine the effects of features of different sizes that are captured in layers of different depths of the network.
		\item As CIFAR 10 consists of naturally occurring objects they exhibit a high degree of intra class spatial variance. In other words, the class specific template has a high variance. Hence, combining multiple instances of whole models performs the best. The concepts required for classification are too abstract to be extracted from shallower depths.
	\end{enumerate}
	\item Finally from the McNemar test conducted between the strong and mean inference methods as well as their comparison with standard networks like ResNet18 or GoogLeNet we can draw some conclusions. The p-values of the McNemar test are provided in table \ref{tab:mcnemar}. Among the four proposed model we can observe that other than M1, all the other models are showing significant difference in the learned concepts. Among the datasets only MNIST dataset didn't demonstrate statistical significance. This is probably due to the saturation in the performance for MNIST dataset.
\end{enumerate}

\begin{table*}[]
	\centering
	\caption{McNemar Test between various proposed methods and existing networks to demonstrate significant difference among the classifiers (p\textless{}0.05 suggests that difference is significant)}
	\label{tab:mcnemar}
	\resizebox{\textwidth}{!}{%
		\begin{tabular}{lllllllll}
			\Xhline{2pt}
			\multicolumn{1}{c}{\textbf{Datasets}} & \multicolumn{1}{c}{\textbf{\begin{tabular}[c]{@{}c@{}}ResNet18\\ vs\\ M1 Mean\end{tabular}}} & \multicolumn{1}{c}{\textbf{\begin{tabular}[c]{@{}c@{}}ResNet18\\ vs\\ M1 Strong\end{tabular}}} & \multicolumn{1}{c}{\textbf{\begin{tabular}[c]{@{}c@{}}M1 Mean\\ vs\\ M1 Strong\end{tabular}}} & \multicolumn{1}{c}{\textbf{\begin{tabular}[c]{@{}c@{}}GoogleLeNet\\ vs \\ M2 Mean\end{tabular}}} & \multicolumn{1}{c}{\textbf{\begin{tabular}[c]{@{}c@{}}GoogleLeNet\\ vs \\ M2 Strong\end{tabular}}} & \multicolumn{1}{c}{\textbf{\begin{tabular}[c]{@{}c@{}}M2 Strong\\ vs \\ M2 Mean\end{tabular}}} & \multicolumn{1}{c}{\textbf{\begin{tabular}[c]{@{}c@{}}M3 Mean\\ vs\\ M3 Strong\end{tabular}}} & \multicolumn{1}{c}{\textbf{\begin{tabular}[c]{@{}c@{}}M4 Mean\\ vs\\ M4 Strong\end{tabular}}} \\ \hline
			\textit{\textbf{CMATERdb 3.1.1}}      & {\color[HTML]{FF0000} \textbf{0.2467}}                                                          & {\color[HTML]{FF0000} \textbf{0.2344}}                                                            & {\color[HTML]{FF0000} \textbf{0.1274}}                                                           & {\color[HTML]{00B050} \textbf{0.0126}}                                                           & {\color[HTML]{00B050} \textbf{0.0084}}                                                             & {\color[HTML]{00B050} \textbf{0.0973}}                                                         & {\color[HTML]{00B050} \textbf{0.0071}}                                                        & {\color[HTML]{00B050} \textbf{0.0017}}                                                        \\
			\textit{\textbf{CMATERdb 3.2.1}}      & {\color[HTML]{FF0000} \textbf{0.2253}}                                                          & {\color[HTML]{FF0000} \textbf{0.1742}}                                                            & {\color[HTML]{FF0000} \textbf{0.1364}}                                                           & {\color[HTML]{00B050} \textbf{0.0095}}                                                           & {\color[HTML]{00B050} \textbf{0.0077}}                                                             & {\color[HTML]{FF0000} \textbf{0.1012}}                                                            & {\color[HTML]{00B050} \textbf{0.0003}}                                                        & {\color[HTML]{00B050} \textbf{0.0010}}                                                        \\
			\textit{\textbf{CMATERdb 3.3.1}}      & {\color[HTML]{FF0000} \textbf{0.1516}}                                                          & {\color[HTML]{FF0000} \textbf{0.1056}}                                                            & {\color[HTML]{FF0000} \textbf{0.0912}}                                                           & {\color[HTML]{00B050} \textbf{0.0110}}                                                           & {\color[HTML]{00B050} \textbf{0.0108}}                                                             & {\color[HTML]{FF0000} \textbf{0.3113}}                                                            & {\color[HTML]{00B050} \textbf{0.0014}}                                                        & {\color[HTML]{00B050} \textbf{0.0011}}                                                        \\
			\textit{\textbf{CMATERdb 3.4.1}}      & {\color[HTML]{FF0000} \textbf{0.2833}}                                                          & {\color[HTML]{FF0000} \textbf{0.2036}}                                                            & {\color[HTML]{FF0000} \textbf{0.0508}}                                                           & {\color[HTML]{00B050} \textbf{0.0129}}                                                           & {\color[HTML]{00B050} \textbf{0.0115}}                                                             & {\color[HTML]{FF0000} \textbf{0.0884}}                                                            & {\color[HTML]{00B050} \textbf{0.0012}}                                                        & {\color[HTML]{00B050} \textbf{0.0022}}                                                        \\
			\textit{\textbf{CMATERdb 3.1.2}}      & {\color[HTML]{FF0000} \textbf{0.1139}}                                                          & {\color[HTML]{00B050} \textbf{0.0451}}                                                         & {\color[HTML]{00B050} \textbf{0.0000}}                                                        & {\color[HTML]{00B050} \textbf{0.0154}}                                                           & {\color[HTML]{00B050} \textbf{0.0118}}                                                             & {\color[HTML]{00B050} \textbf{0.0146}}                                                         & {\color[HTML]{FF0000} \textbf{0.0528}}                                                           & {\color[HTML]{00B050} \textbf{0.0088}}                                                        \\
			\textit{\textbf{CMATERdb 3.1.3}}      & {\color[HTML]{FF0000} \textbf{0.1087}}                                                          & {\color[HTML]{00B050} \textbf{0.0078}}                                                         & {\color[HTML]{00B050} \textbf{0.0000}}                                                        & {\color[HTML]{00B050} \textbf{0.0176}}                                                           & {\color[HTML]{00B050} \textbf{0.0091}}                                                             & {\color[HTML]{00B050} \textbf{0.0133}}                                                         & {\color[HTML]{00B050} \textbf{0.0096}}                                                        & {\color[HTML]{00B050} \textbf{0.0000}}                                                        \\
			\textit{\textbf{MNIST}}               & {\color[HTML]{FF0000} \textbf{0.4772}}                                                          & {\color[HTML]{FF0000} \textbf{0.5987}}                                                            & {\color[HTML]{FF0000} \textbf{0.3726}}                                                           & {\color[HTML]{FF0000} \textbf{0.5630}}                                                              & {\color[HTML]{FF0000} \textbf{0.3753}}                                                                & {\color[HTML]{FF0000} \textbf{0.4510}}                                                            & {\color[HTML]{FF0000} \textbf{0.1368}}                                                           & {\color[HTML]{FF0000} \textbf{0.3147}}                                                           \\
			\textit{\textbf{Fashion-MNIST}}       & {\color[HTML]{FF0000} \textbf{0.1638}}                                                          & {\color[HTML]{FF0000} \textbf{0.1065}}                                                            & {\color[HTML]{00B050} \textbf{0.0546}}                                                        & {\color[HTML]{FF0000} \textbf{0.1464}}                                                              & {\color[HTML]{00B050} \textbf{0.0000}}                                                             & {\color[HTML]{FF0000} \textbf{0.1724}}                                                            & {\color[HTML]{00B050} \textbf{0.0012}}                                                        & {\color[HTML]{00B050} \textbf{0.0000}}                                                        \\
			\textit{\textbf{NotMNIST}}            & {\color[HTML]{FF0000} \textbf{0.1716}}                                                          & {\color[HTML]{FF0000} \textbf{0.1013}}                                                            & {\color[HTML]{FF0000} \textbf{0.1142}}                                                           & {\color[HTML]{FF0000} \textbf{0.1954}}                                                              & {\color[HTML]{00B050} \textbf{0.0001}}                                                             & {\color[HTML]{00B050} \textbf{0.0430}}                                                         & {\color[HTML]{00B050} \textbf{0.0437}}                                                        & {\color[HTML]{00B050} \textbf{0.0000}}                                                        \\
			\textit{\textbf{Kannada-MNIST}}       & {\color[HTML]{FF0000} \textbf{0.2009}}                                                          & {\color[HTML]{00B050} \textbf{0.0007}}                                                         & {\color[HTML]{00B050} \textbf{0.0062}}                                                        & {\color[HTML]{FF0000} \textbf{0.1735}}                                                              & {\color[HTML]{00B050} \textbf{0.0320}}                                                             & {\color[HTML]{00B050} \textbf{0.0098}}                                                         & {\color[HTML]{00B050} \textbf{0.0000}}                                                        & {\color[HTML]{00B050} \textbf{0.0426}}                                                        \\
			\textit{\textbf{Kuzushiji-MNIST}}     & {\color[HTML]{FF0000} \textbf{0.1863}}                                                          & {\color[HTML]{FF0000} \textbf{0.1139}}                                                            & {\color[HTML]{FF0000} \textbf{0.2366}}                                                           & {\color[HTML]{FF0000} \textbf{0.2573}}                                                              & {\color[HTML]{00B050} \textbf{0.0175}}                                                             & {\color[HTML]{00B050} \textbf{0.0226}}                                                         & {\color[HTML]{00B050} \textbf{0.0000}}                                                        & {\color[HTML]{00B050} \textbf{0.0387}}                                                        \\
			\textit{\textbf{SVHN}}                & {\color[HTML]{FF0000} \textbf{0.1039}}                                                          & {\color[HTML]{00B050} \textbf{0.0023}}                                                         & {\color[HTML]{00B050} \textbf{0.0040}}                                                        & {\color[HTML]{FF0000} \textbf{0.0956}}                                                              & {\color[HTML]{00B050} \textbf{0.0001}}                                                             & {\color[HTML]{00B050} \textbf{0.0001}}                                                         & {\color[HTML]{00B050} \textbf{0.0000}}                                                        & {\color[HTML]{00B050} \textbf{0.0067}}                                                        \\
			\textit{\textbf{Cifar10}}             & {\color[HTML]{00B050} \textbf{0.0064}}                                                       & {\color[HTML]{FF0000} \textbf{0.0964}}                                                            & {\color[HTML]{FF0000} \textbf{0.0736}}                                                           & {\color[HTML]{FF0000} \textbf{0.1007}}                                                              & {\color[HTML]{00B050} \textbf{0.0411}}                                                             & {\color[HTML]{00B050} \textbf{0.0012}}                                                         & {\color[HTML]{00B050} \textbf{0.0141}}                                                        & {\color[HTML]{FF0000} \textbf{0.0785}}                                                           \\ \Xhline{2pt}
		\end{tabular}%
	}
\end{table*}

\section{Conclusion}
In this work, we have proposed a viable alternative to traditional averaging techniques for aggregating logits from multiple columns. The proposed approach analyzes weakly learnt concepts and guides the backpropagation gradients along those pathways to address the weaknesses. Two inference protocols have been proposed that can either focus on the strongest pathway or even average of the improved logits. The proposed model triumphs of the standard averaging technique across 13 different datasets. It evens achieves new benchmark scores in a few datasets. It is also established statistically that for several applications, the learned concepts of the proposed method are significantly different than the competing methods. Currently, the proposed method works only in the output layer with logits, however it may be possible to adopt a similar approach in future works for more inner layers.
\newpage
\bibliographystyle{unsrt}
\bibliography{ref}
\end{document}